\documentclass[runningheads]{llncs}
\usepackage[T1]{fontenc}
\usepackage{graphicx}
\usepackage{color}
\usepackage{hyperref}

\usepackage{stmaryrd}
\usepackage{amsmath}
\usepackage{amssymb}
\usepackage{mathtools}
\usepackage{amsthm}
\usepackage{pgfplots}
\usepackage{multirow}
\usepackage{array}
\usepackage{listings}
\usepackage{inconsolata}
\usepackage{graphicx}
\usepackage{xcolor}         
\newcommand{\gray}[1]{\textcolor{gray}{#1}}
\newcommand{\green}[1]{\textcolor{green!50!black}{#1}}

\usepackage{subcaption}
\usepackage{booktabs}
\usepackage{cite}

\newcommand{\fref}[1]{Fig.~\ref{#1}}
\newcommand{\sref}[1]{Section~\ref{#1}}
\newcommand{\cref}[1]{Chapter~\ref{#1}}

\newcommand{\tref}[1]{Table~\ref{#1}}
\newcommand{\eref}[1]{Eq.~\ref{#1}}

\pgfplotsset{compat=1.18}

\begin{document}
%
\title{Channel-Aware Probing for Multi-Channel Imaging}
%
%

\author{Umar Marikkar\inst{1} \and
Syed Sameed Husain\inst{1} \and
Muhammad Awais\inst{1} \and
Sara Atito\inst{1,2}}
\authorrunning{U. Marikkar et al.}
%
\institute{Institute for People-Centered AI, University of Surrey \and
Centre of Vision, Speech and Signal Processing, University of Surrey
\email{\{u.marikkar,sameed.husain,muhammad.awais,sara.atito\}@surrey.ac.uk}}
\maketitle              
\begin{abstract}

Training and evaluating vision encoders on Multi-Channel Imaging (MCI) data remains challenging as channel configurations vary across datasets, preventing fixed-channel training and limiting reuse of pre-trained encoders on new channel settings. Prior work trains MCI encoders but typically evaluates them via full fine-tuning, leaving probing with frozen pre-trained encoders comparatively underexplored. Existing studies that perform probing largely focus on improving representations, rather than how to best leverage fixed representations for downstream tasks. Although the latter problem has been studied in other domains, directly transferring those strategies to MCI yields weak results, even worse than training from scratch. 
We therefore propose Channel-Aware Probing (CAP), which exploits the intrinsic inter-channel diversity in MCI datasets by controlling feature flow at both the encoder and probe levels. CAP uses Independent Feature Encoding (IFE) to encode each channel separately, and Decoupled Pooling (DCP) to pool within channels before aggregating across channels. 
Across three MCI benchmarks, CAP consistently improves probing performance over the default probing protocol, matches fine-tuning from scratch, and largely reduces the gap to full fine-tuning from the same MCI pre-trained checkpoints. Code can be found in \url{https://github.com/umarikkar/CAP}

\end{abstract}

\section{Introduction}

Training and evaluating vision encoders on Multi-Channel Imaging (MCI) data remains challenging due to differences in channel configurations across datasets. For example, the CHAMMI multi-channel microscopy imaging benchmark \cite{chen2023chammi} contains images with different channel counts, spanning three immunofluorescent imaging datasets \cite{thul2017subcellular,chandrasekaran2023jump,viana2023integrated}. Other satellite imaging datasets may differ by channel counts depending on the sensor types and whether multi-modal sources are fused \cite{so2sat}. This heterogeneity of channels prevents common fixed-channel training and evaluation in computer vision, and also prevents the direct insertion of MCI pre-trained vision encoders on new datasets with novel channel configurations. 



\begin{figure*}[t]
\centering

\begin{minipage}[c]{0.35\linewidth}
    \centering
    \includegraphics[width=\linewidth]{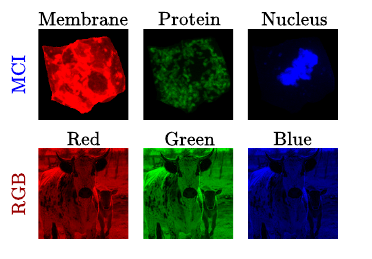}
    \subcaption{}
    \label{fig:mci_rgb}
\end{minipage}%
\begin{minipage}[c]{0.64\linewidth}
    \centering
    \includegraphics[width=\linewidth]{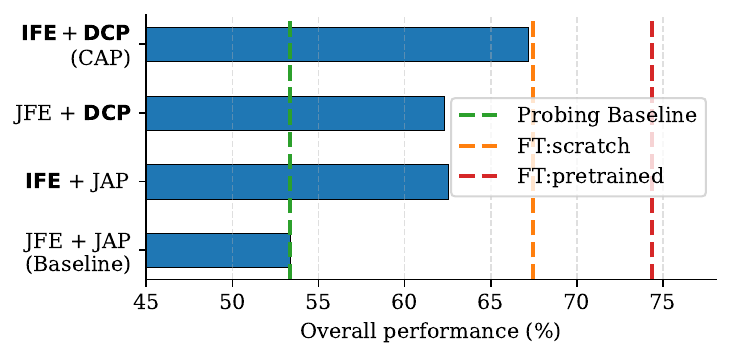}
    \subcaption{}
    \label{fig:intro_results}
\end{minipage}

\caption{(a) Channel diversity in Multi-Channel Imaging (MCI) datasets, where individual channels carry distinct semantic information. The MCI sample is obtained from WTC-11 \cite{viana2023integrated} and the RGB sample from Pascal-VOC \cite{everingham2010pascal}. (b) Comparison to probing baseline and full fine-tuning baselines. \{J/I\}-FE: \{Joint/\textbf{Independent}\}-Feature Encoding. \{JA/DC\}-P: \{Joint-Attentive/\textbf{Decoupled}\}-Pooling. We plot the results for the \(\mathtt{protobin}\) pooling architecture for each setting.}
\label{fig:interchan}
\end{figure*}


To address this, prior work proposes Multi-Channel Vision Transformers (MC-ViTs) for MCI data, where the gold standard is to pre-train a model on individual channel images using self-supervised representation learning, and fine-tune in a multi-channel setting \cite{lian2025isolated,xun2023microsnoop}. While some of these studies also perform probing evaluations, where the pre-trained encoder is frozen and a lightweight probing network is trained on fixed feature representations, they primarily focus on improving representation quality rather than on how these fixed representations are leveraged for downstream tasks.


This leaves an open research gap in how to effectively leverage fixed representations obtained from MCI pre-trained models for downstream tasks. This problem has been actively studied in other domains such as general computer vision \cite{psomas2023keep,assran2025v,przewikezlikowski2025beyond,psomas2025attention}, audio \cite{rauch2025can,rauch2025unmute}, and histopathology \cite{hense2024xmil,shao2025mil}. However, our preliminary experiments show that directly applying these methods to MCI data performs substantially worse than full fine-tuning, with a much larger gap observed compared to probing vs. fine-tuning in other domains, and even worse than fine-tuning from scratch. Therefore, applying existing probing approaches with frozen encoders is unrealistic for MCI data, thereby relying on full fine-tuning. This further motivates effective probing strategies for MCI datasets.

We begin with the knowledge that unlike in natural images, individual MCI channels encode distinct information, as shown in \fref{fig:mci_rgb} and verified in literature \cite{thul2017subcellular,chandrasekaran2023jump,viana2023integrated,so2sat}. Hence, individual channels are statistically diverse. Motivated by this, we propose \textbf{Channel-Aware Probing (CAP)}, a probing framework that improves over the standard probing protocol of MC-ViTs by exploiting inter-channel diversity in MCI data. Specifically, CAP exploits this by controlling feature flow at two stages: at the encoder level, by governing how fixed features are pre-computed, and at the probe level, by determining how these features are aggregated and processed by the probing network.

Standard probing in MC-ViTs exhibits issues at two levels. (1) Encoder-level: the standard inference protocol uses Joint Feature Encoding (JFE),  where all channels are encoded jointly in a single sequence. This joint encoding can dilute inter-channel diversity and weaken channel-specific representations. (2) Probe-level: existing probes typically apply Joint Attentive Pooling (JAP) in a channel-agnostic manner, which can obscure salient signals from less dominant channels. 

We address these limitations with two corresponding design changes. (1) For the encoder, we use Independent Feature Encoding (IFE), encoding each channel as its own sequence to preserve channel-specific information. (2) For the probe, we introduce Decoupled Pooling (DCP), a hierarchical strategy that first aggregates features within each channel and then pools across channels, preserving channel-level saliency while enabling cross-channel integration.

Across standard MCI benchmarks \cite{chen2023chammi,chandrasekaran2024three,so2sat}, {Independent Feature Encoding (IFE)} alone increases input feature diversity to the probe and yields a {+9\%} improvement over {Joint Feature Encoding (JFE)} (\fref{fig:intro_results}). At the probe level, {Decoupled Pooling (DCP)} alone improves performance by {+9\%} compared to {Joint Attentive Pooling (JAP)}. Combined, {Channel-Aware Probing (CAP)} achieves a {14\%} improvement over the standard probing baseline (JFE+JAP). More importantly, CAP matches the performance of full fine-tuning from scratch, and recovers approximately two thirds of the performance gap to full fine-tuning from the same pre-trained encoder, reducing the remaining gap to {7\%}, whereas the standard probing baseline remains {21\%} behind. These results indicate that explicitly leveraging inherent channel diversity is an effective step toward efficient training on MCI datasets without relying on full fine-tuning.

\section{Related work}

This work investigates leveraging pre-trained MC-ViTs for downstream tasks using probing. Therefore, we summarize current works on MC-ViTs, and works on probing architectures proposed in other domains. 

\subsection{Multi-Channel Vision Transformers (MC-VITs)}

Vision encoders for handling MCI data have been proposed for climate modeling \cite{9672063,nguyen2023climax}, cell microscopy \cite{chen2023chammi} and satellite imagery \cite{so2sat}. The recent studies on encoders for MCI data fall under MC-ViTs, where specialized methods are proposed to train Vision Transformers (ViTs) \cite{dosovitskiy2020image} effectively on MCI data. 

In comparison to vanilla ViTs which embed spatial patches of different channels into a single data token, the standard implementation of MC-ViTs embed spatial patches of individual channels yield their own token embeddings with spatially coherent positional embeddings. This is first seen in ChAdaViT \cite{bourriez2024chada}, which also proposes learnable channel embeddings which are added to the token embeddings. ChannelViT, DiChaViT and ChaMAEVIT  \cite{bao2023channel,pham2024enhancing,pham2025cha} build on this by incorporating different training strategies. However, these methods are proposed for full fine-tuning from scratch, where the encoder weights are guided by a task-specific loss. 

Conversely, IC-ViT \cite{lian2025isolated} proposes pre-training channel-agnostic ViTs prior to fine-tuning, by pre-training models on individual channel images. They show that single-channel pre-training and then performing fine-tuning with the basic MC-ViT framework improves downstream performance even in a multi-channel setting. However, IC-ViT does not evaluate under frozen-encoder settings, a standard protocol for assessing pre-trained representations. There is existing work evaluating frozen encoders on MCI downstream tasks. SubCell \cite{gupta2024subcell} and Dino4Cells \cite{doron2023unbiased} pre-train vanilla ViTs on multi-channel microscopy data and perform downstream experiments using probing; however, they do not fall under MC-ViTs as they are not channel-adaptive. 

The default MC-ViT inference protocol during evaluation is to encode a long concatenated sequence of all tokenized embeddings of all channels, which we name Joint Feature Encoding (JFE). In contrast, we investigate Independent Feature Encoding (IFE) to preserve channel diversity, and incorporate this into the probing pipeline. 



\subsection{Probing Architectures}

As existing encoders on MCI data (MC-ViTs) are variants of ViTs, we focus on probing studies performed on ViTs. Recent works propose probing strategies that aggregate all token features from ViTs into a single image-level representation, rather than relying solely on the latent \(\mathtt{[cls]}\) token of the encoder. For example, AbMILp \cite{przewikezlikowski2025beyond} demonstrates the limitation of \(\mathtt{[cls]}\) tokens in ViTs and attributes this limitation to pre-training strategies. Therefore they adopt a Multiple Instance Learning (MIL) formulation, treating patch tokens as instances in a bag and aggregating them into an image-level embedding. Similarly, V-JEPA \cite{assran2025v} introduces attentive probe training using multi-head cross-attention with a latent query token over patch tokens. Subsequent works propose similar attention-based pooling mechanisms, primarily targeting improved efficiency by simplifying or removing projection heads \cite{psomas2025attention,psomas2023keep}.

Beyond general computer vision, MIL-style attentive pooling is standard in histopathology, where whole slide images are prohibitively large for full encoder training \cite{hense2024xmil,shao2025mil}. In audio, studies propose  pooling time-frequency tokens using prototypical probes, where a set of learnable probes attend to feature vectors \cite{rauch2025can,rauch2025unmute}. For MCI data, Cha-MAE-ViT \cite{pham2025cha} proposes hybrid token fusion, pooling \(\mathtt{[cls]}\), patch, and auxiliary tokens via a Set-Transformer MAB block \cite{lee2019set}.

We categorize the above probing methods proposed in literature as Joint Attentive Pooling (JAP), where a learned aggregation is performed via an arbitrary pooling function, on all feature vectors representative of a given image. However, we note that none of these pooling approaches explicitly model or exploit channel heterogeneity. We are the first to investigate channel-aware structure into the probing pipeline via Decoupled Pooling (DCP).

\section{Methodology}

The methodology is structured as follows. First, we explore the inherent differences between MCI and fixed-channel datasets in \sref{sec:motivation}, and show inter-channel diversity as a distinct property in MCI datasets. This is the primary factor that motivates CAP. We then describe the methodology of CAP in \sref{cap}. 

\subsection{Exploring distinct properties in MCI datasets}
\label{sec:motivation}

We first explore how the information is carried in individual channels in MCI datasets versus natural (fixed-channel) image datasets. To do this, we analyze feature diversity between channels. Specifically, given an MCI dataset, we measure the feature cosine similarity between channels within a given instance. A high similarity between channels within an instance implies that there is lesser distinct information available from individual channels, and vise versa.

\begin{figure*}[t]
\centering
    \centering
        \includegraphics[width=0.5\linewidth]{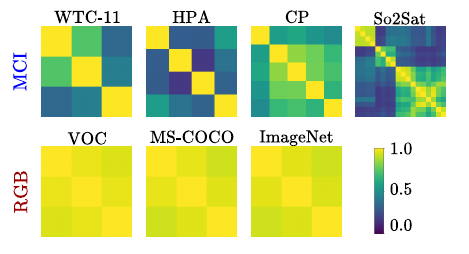}
    \caption{Comparison of inter-channel feature diversity between MCI and RGB datasets. We compute the cosine similarity between \(\mathtt{[cls]}\) tokens of individual channel features within a given instance, averaged over 1000 random instances.}
    \label{fig:interchan_a}
\end{figure*}

For a given dataset, we perform this experiment by passing individual channel images through a pre-trained encoder, to obtain a single feature (\(\mathtt{[cls]}\) token) per channel. We then compute the cosine similarity of this \(\mathtt{[cls]}\) token with other channel \(\mathtt{[cls]}\) tokens within the same instance. We perform this experiment on WTC-11, HPA and CP datasets (in the CHAMMI benchmark), and So2Sat. We compare the property of inter-channel diversity with instances from Pascal-VOC, MS-COCO and ImageNet. We use a single-channel pre-trained ViT-S for MCI datasets \cite{lian2025isolated}, and DINOv2 pre-trained ViT-S \cite{oquab2023dinov2} for RGB datasets. 

The computed average cosine similarities between channels within an instance is shown in \fref{fig:interchan_a}. Here, we observe a clear distinction between the MCI (blue) and RGB (red) groups, where the per-channel cosine similarity between each of the channels in RGB datasets is closer to 1, implying that individual channels share similar semantic information. In contrast, for MCI datasets, we observe a high degree of discrimination (\(\sim\) 0.5 cosine similarity) between different channel features of the same instance. This builds the motivation towards channel-aware probing, where in MCI datasets, each of the channels within a given instance contains distinct information, and therefore should be viewed independently.

\subsection{Channel-Aware Probing for Multi-Channel Imaging}
\label{cap}

Based on the observed property of inter-channel diversity in MCI datasets, we propose Channel-Aware Probing (CAP) to perform effective probing evaluations in MCI datasets. We first describe Independent Feature Encoding (IFE) for feature extraction in contrast to Joint Feature Encoding (JFE) of channels which is the standard in existing MC-ViTs. Next, we describe Decoupled Pooling (DCP), an improvement on Joint Attentive Pooling (JAP) for MCI datasets, that preserves channel importance during probing. 

\subsubsection{Maximizing feature diversity via Independent Feature Encoding.}

The default encoding protocol of MC-ViTs during fine-tuning and inference is to obtain a joint, channel-conditioned representation of the input image. We name this Joint Feature Encoding (JFE). 

\begin{figure*}[t]
\centering

\begin{subfigure}[t]{0.2\linewidth}
    \centering
    \includegraphics[height=3cm]{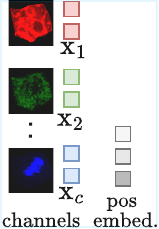}
\end{subfigure}
\hfill
\begin{subfigure}[t]{0.35\linewidth}
    \centering
    \includegraphics[height=3cm]{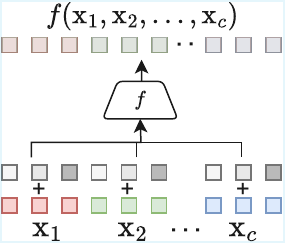}  
    \caption{}
    \label{fig:feats_joint}
\end{subfigure}
\hfill
\begin{subfigure}[t]{0.38\linewidth}
    \centering
    \includegraphics[height=3cm]{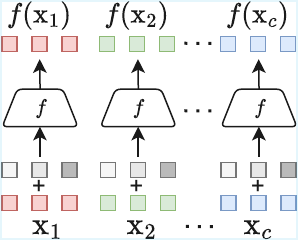}    
    \caption{}
    \label{fig:feats_sep}
\end{subfigure}
\caption{(a) Joint feature encoding, the default encoding protocol in MC-ViTs. Information is shared between channels, resulting in pre-conditioned output features. (b) Independent feature encoding, where information is not shared between channels in the computation of the overall feature map. \(\mathrm{x}_i\) denotes the set of tokenized embeddings for channel \(i\).}
\label{test}
\end{figure*}

Following general nomenclature of ViTs, MC-ViTs embed patches of individual channels into data tokens, add positional embeddings where the positional embeddings are shared between channels for a set position, and then concatenate all data tokens along the sequence dimension. These data tokens are then passed to the encoder blocks, which perform self-attention between all data tokens similar to vanilla ViTs. The schematic of JFE is shown in \fref{fig:feats_joint}. 

We formalize the JFE as follows. Given an input set of tokenized patch embeddings \(\mathbf{x} = \{\mathrm{x}_1, \mathrm{x}_2, \dots ,\mathrm{x}_c\} \in \mathbb{R}^{C \times N \times D}\) and positional embeddings \(\mathrm{x}_\text{pos} \in \mathbb{R}^{N \times D}\), we write JFE as,
\begin{align}
\label{eq:pos}
    \hat{\mathbf{x}} &= \{\mathrm{x}_1 \!+ \!\mathrm{x}_\text{pos}, \mathrm{x}_2 \!+ \!\mathrm{x}_\text{pos}, \dots , \mathrm{x}_c \!+ \!\mathrm{x}_\text{pos}\} \in  \mathbb{R}^{C\times N \times D},  \\
    \hat{\mathbf{x}}_\text{flat} &= \text{concat}(\hat{\mathbf{x}}, \dim\!=\!C) \in  \mathbb{R}^{CN \times D}, \\
    \label{eq:jfe}
    X_\text{JFE} &= f( \hat{\mathbf{x}}_\text{flat}) \in \mathbb{R}^{CN \times D},
\end{align}
where \(f(\cdot)\) denotes the sequence of encoder blocks, \(C\) is the total number of channels, \(N\) is the number of feature vectors per channel, and \(D\) is the output feature dimensionality of the encoder.  In \eref{eq:jfe}, we note that each token of a given channel accumulates information from all tokens of all channels. We suspect that this channel mixing hinders the diversity of output features. 

To address this, we implement {Independent Feature Encoding} (IFE), where we view channels independently during the encoding step, as shown in \fref{fig:feats_sep}. Unlike JFE, IFE does not concatenate the set of token embeddings prior to passing to the encoder. Rather, each set of tokens per channel is encoded independently. Given the set of token embeddings \(\hat{\mathbf{x}}\) (with added positional embeddings using \eref{eq:pos}), we write IFE as, 
\begin{align}
    \label{eq:ife}
    X_\text{IFE} &= \{f( \hat{\mathrm{x}}_1), f( \hat{\mathrm{x}}_2), \dots, f( \hat{\mathrm{x}}_c)\} \in \mathbb{R}^{C \times N \times D}. 
\end{align}

We show later in \sref{sec:analysis} that this simple restriction of information sharing between channels using IFE yields a greater diversity in input features to the probe in comparison to JFE, resulting in better downstream metrics.

\subsubsection{Leveraging diverse channel features via Decoupled Pooling.} Here, we outline the proposed pooling framework for per-computed MCI features. We first describe joint pooling; the standard approach of pooling features in general computer vision, and then outline the proposed decoupled pooling strategy. 

We begin by defining the feature input to the probing network. Fixed feature representations for a multi-channel image can be obtained either by independent encoding of channels as in \fref{fig:feats_joint}, or joint encoding of channels as in \fref{fig:feats_sep}, through an encoder \(f(\cdot)\). Based on the previous finding that independent encoding yields higher feature diversity, we consider independent encoding of channels as the input to the probing network. Hence, for a given MCI image, we define the fixed multi-channel feature representation as,  
\begin{align}
    X \in \mathbb{R}^{C \times N\times D},
\end{align}
where \(C\) is the number of channels, \(N\) is the number of features (tokens) per channel, and \(D\) is the feature dimensionality of the encoder. Next, we define a pooling function \(g\),
\begin{align}
    g:\mathbb{R}^{M \times D} \mapsto \mathbb{R}^{D},
\end{align}
that maps a set of \(M\) input feature vectors of dimensionality \(D\) into a single representative vector. This function may represent any of the pooling strategies proposed in the literature \cite{psomas2023keep,assran2025v,przewikezlikowski2025beyond,psomas2025attention}. Our focus lies not on the architecture of \(g(\cdot)\), but on how to control the information flow of the input \(X\) through \(g(\cdot)\) using the notion of channel diversity in MCI data to maximize downstream performance. 

\begin{figure}[t]
\centering
\begin{minipage}[c]{0.34\linewidth}
    \centering
        \begin{subfigure}[b]{\linewidth}
        \centering
        \includegraphics[width=\linewidth]{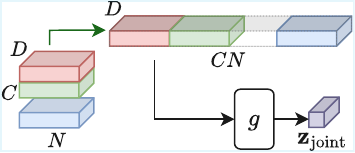}
        \subcaption{}
        \label{fig:probing_a}
    \end{subfigure}
\end{minipage}%
\hfill
\begin{minipage}[c]{0.54\linewidth}
    \centering
    \begin{subfigure}[t]{\linewidth}
        \centering
        \includegraphics[width=\linewidth]{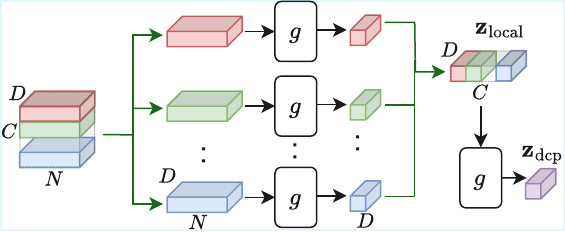}
        \subcaption{}
        \label{fig:probing_b}
    \end{subfigure}
\end{minipage}
\caption{(a) JAP: All channel-wise features are concatenated into a single sequence and then pooled. (b) DCP: Each set of channel-wise features is pooled independently to yield a set of single feature vectors per channel, which are pooled again via same function. \green{green arrows \(\rightarrow\)}: reshaping operations.}
\label{test2}
\end{figure}

We first describe {Joint Attentive Pooling} (JAP). In general probing, all feature vectors are treated jointly as a single global set to be aggregated by the pooling architecture \(g(\cdot)\). Applied to multi-channel features, this corresponds to concatenating all channel features along the sequence dimension and aggregating them in a single step, as illustrated in \fref{fig:probing_a}. For the input \(X \in \mathbb{R}^{C \times N \times D}\), we write joint pooling \(g_\text{joint}\) as
\begin{align}
    \mathbf{X}_\text{joint} &= \text{concat}(X_1, X_2, \dots, X_C) \in \mathbb{R}^{(C N) \times D}, \\
    \mathbf{z}_\text{joint} &= g\big(\mathbf{X}_{j}\big) := g_\text{joint}(X)  \in \mathbb{R}^{D},
\end{align}

where \(X_c \in \mathbb{R}^{N \times D}\) denotes the feature set of channel \(c\), and \(g(\cdot)\) is the base pooling function applied over a set of feature vectors. Joint pooling identifies the most relevant features for a downstream task in a channel agnostic manner. In contrast, pooling features within each channel first can help preserve channel-specific salient information. A subsequent pooling across these per-channel representations may capture information from all channels that could be overlooked by a single global pooling. 

Hence, we propose {Decoupled Pooling} (DCP), which first pools feature vectors spatially, and then channel-wise through the same function \(g(\cdot)\), as illustrated in \fref{fig:probing_b}. We write decoupled pooling \(g_\text{dcp}\) as
\begin{align}
    \mathbf{z}_{\text{local}} &= \big( g(X_1), g(X_2), \dots, g(X_C) \big) \in \mathbb{R}^{C \times D}, \\
    \label{dcp_fwd}
    \mathbf{z}_{\text{dcp}} &= g\big(\mathbf{z}_{\mathtt{local}}\big) := g_{\text{dcp}}(X) \in \mathbb{R}^{D},
\end{align}

where \(X_c \in \mathbb{R}^{N \times D}\) denotes the feature set of channel \(c\), and \(g(\cdot)\) is the base aggregation function. This hierarchical design encourages the probing network to preserve channel‑specific characteristics before combining them into a unified representation. The overall parameter count of \(g_\text{dcp}\)  remains the same as \(g_\text{joint}\) due to the shared pooling function. Although \(g_\text{dcp}\) contains two sequential forward passes of \(g(\cdot)\), the second pass through \(g(\cdot)\) contains a sequence length of \(C\) which is the number of channels. Hence, the resulting increase FLOPs is negligible as \(C \ll N\), which we show later in \sref{sec:compute}. 

The combination of IFE in \fref{fig:feats_sep} and DCP in \fref{fig:probing_b} leverages the inherent channel diversity in multi-channel images, resulting in distinct feature learning as opposed to joint encoding of channels and joint pooling afterwards.



\subsubsection{Datasets and Tasks.} We evaluate our method on three benchmarks typically used in existing MC-ViT studies; CHAMMI, JUMP-CP, and So2Sat. We summarize them below.

\emph{CHAMMI} \cite{chen2023chammi} is a microscopy imaging benchmark containing data from WTC-11 \cite{viana2023integrated}, HPA \cite{hpa-single-cell-image-classification}, and CP \cite{bray2016cell} datasets, with each dataset containing images with 3, 4 and 5 fluorescent channels respectively. The combined dataset contains 220K total and 100K training images. The benchmark contains of 9 downstream tasks, of which 6 are out-of-distribution (OOD) tasks. Following existing MC-ViT studies (IC-ViT, DiChaViT etc.), we report the OOD average on WTC-11 and HPA. Per-task results can be found in Appendix A.2. 

The \emph{JUMP-CP} \cite{chandrasekaran2024three} dataset contains 8-channel images (5 fluorescent, 3 bright-field) of cells that have been perturbed with chemical compounds. The dataset forms 127K training, 45K validation and 45K test images. The downstream task is solved as a multi-class classification problem, with 161 classes that represent 160 unique drug perturbations and a single control treatment. 

\emph{So2Sat} \cite{so2sat} benchmark is evaluated on a remote sensing dataset with satellite imagery data combined from Sentinel-1 (8 radar channels) and Sentinel-2 (10 multi-spectral channels). So2Sat contains 352K training, 24K validation and 24K test images. The downstream task is solved as a 17-class classification problem, with each class representing a distinct surface property such as vegetation, paving or buildings. 

\subsubsection{Baselines and Comparisons.} The primary contribution of this work is Channel-Aware Probing (CAP), which  controls the information flow of fixed MCI feature representations when performing probing. Therefore, our main comparison lies against the existing probing protocol in other domains which consists of Joint Feature Encoding (JFE) and Joint Attentive Pooling (JAP). 

To verify robustness the pooling architecture, we implement 6 architectures published in recent works: $\mathtt{simpool}$\cite{psomas2023keep} and $\mathtt{abmilp}$\cite{przewikezlikowski2025beyond} which perform simple weighted aggregation of feature vectors , $\mathtt{ep}$\cite{psomas2025attention}, $\mathtt{mab}$\cite{lee2019set} and $\mathtt{mhca}$\cite{assran2025v} which perform weighted aggregation with additional linear projections, and \(\mathtt{protobin}\)\cite{rauch2025unmute}, which performs prototype-based aggregation on the feature vectors.

We obtain fixed representations from the publicly available set of model checkpoints from IC-ViT \cite{lian2025isolated}, and compare with the reported full-fine tuning results in IC-ViT to evaluate performance delta from probing to fine-tuning. To further verify robustness to encoder representations, we perform experiments with fixed representations obtained from OpenPhenom \cite{xun2023microsnoop} and an in-house pre-trained iBOT model on HPA \cite{hpa-single-cell-image-classification,zhou2021ibot}.


\subsubsection{Implementation details.}

For each experiment, which consists of permutations of  \{Dataset\}-\{JFE/IFE\}-\{JAP/DCP\}-\{\(g\)\}, we optimize only the learning rate (LR), as we find that performance is less sensitive to weight decay across \{JFE/IFE\}, \{JAP/DCP\} and \{\(g\)\}. Therefore, we set weight decay to 0.01 for CHAMMI and JUMP-CP, and to 0.1 for So2Sat. We perform 10 initial runs under fixed seed (42) with log-uniform LR range between \(1e^{-5}\) and \(1e^{-2}\), and perform fine search, till the best validation accuracy is obtained at an LR resolution of \(\pm1 e^{p}\), where \(p=\{-5, \dots, -2\}\). We run 4 additional seeds with the best LR and report mean and standard deviation over the 5 total seeds.
We perform each experiment on a RTX 3090 or 2080Ti, with batch size 128.

\begingroup
\setlength{\tabcolsep}{4pt}
\begin{table}[t]
\caption{Downstream performance of CAP vs. (JFE+JAP) on MCI datasets under different pooling architectures (\(g\)). FT:scratch and FT:pretrained have been obtained from DiChaViT \cite{pham2024enhancing} and IC-ViT \cite{lian2025isolated}, respectively. We extract features from IC-ViT pre-trained checkpoints. \(\Delta_\mathtt{CAP}\) indicates improvement over the same pooling architecture.}
\label{tab:lp_results}
\centering
\small
\begin{tabular}{
ll
c @{ }l
c @{ }l
c @{ }l
c @{ }l
c
}
\toprule
Meth. & \(g(\cdot)\)
& \multicolumn{2}{c}{CHAMMI}
& \multicolumn{2}{c}{JUMP}
& \multicolumn{2}{c}{So2Sat}
& \multicolumn{2}{c}{Overall}
& \(\Delta_{\mathtt{CAP}}\) \\
\midrule


& \(\mathtt{simpool}\)
& 44.21 & {\scriptsize$\pm$0.08}
& 26.09 & {\scriptsize$\pm$0.03}
& 62.00 & {\scriptsize$\pm$0.11}
& 44.10 & {\scriptsize$\pm$0.07}
& \\
& \(\mathtt{abmilp}\)
& 52.02 & {\scriptsize$\pm$1.80}
& 31.06 & {\scriptsize$\pm$0.11}
& 62.48 & {\scriptsize$\pm$0.29}
& 48.52 & {\scriptsize$\pm$0.74}
& \\
\(\mathtt{JFE+}\) & \(\mathtt{ep}\)
& 53.92 & {\scriptsize$\pm$0.32}
& 35.02 & {\scriptsize$\pm$0.36}
& 61.83 & {\scriptsize$\pm$0.16}
& 50.26 & {\scriptsize$\pm$0.28}
& \\
\(\mathtt{JAP}\) & \(\mathtt{mab}\)
& 56.03 & {\scriptsize$\pm$1.23}
& 36.19 & {\scriptsize$\pm$0.22}
& 61.58 & {\scriptsize$\pm$0.19}
& 51.26 & {\scriptsize$\pm$0.55}
& \\
& \(\mathtt{mhca}\)
& 54.18 & {\scriptsize$\pm$0.64}
& 34.34 & {\scriptsize$\pm$0.24}
& 61.71 & {\scriptsize$\pm$0.19}
& 50.07 & {\scriptsize$\pm$0.36}
& \\
& \(\mathtt{protobin}\)
& {56.41} & {\scriptsize$\pm$0.92}
& {39.34} & {\scriptsize$\pm$0.24}
& {64.24} & {\scriptsize$\pm$0.05}
& {53.33} & {\scriptsize$\pm$0.40}
& \\

\cmidrule(lr){1-11}

\multirow{6}{*}{\(\mathtt{CAP}\)}
& \(\mathtt{simpool}\)
& 65.16 & {\scriptsize$\pm$1.54}
& 30.38 & {\scriptsize$\pm$0.19}
& 65.34 & {\scriptsize$\pm$0.24}
& 53.63 & {\scriptsize$\pm$0.66}
& \green{$\uparrow$~\small{09.53}} \\
& \(\mathtt{abmilp}\)
& 70.51 & {\scriptsize$\pm$0.43}
& 31.52 & {\scriptsize$\pm$0.62}
& 65.10 & {\scriptsize$\pm$0.33}
& 55.71 & {\scriptsize$\pm$0.46}
& \green{$\uparrow$~\small{07.19}} \\
& \(\mathtt{ep}\)
& 69.68 & {\scriptsize$\pm$0.39}
& 57.85 & {\scriptsize$\pm$0.29}
& 65.92 & {\scriptsize$\pm$0.19}
& 64.48 & {\scriptsize$\pm$0.29}
& \green{$\uparrow$~\small{14.23}} \\
& \(\mathtt{mab}\)
& {70.67} & {\scriptsize$\pm$1.20}
& 60.68 & {\scriptsize$\pm$0.53}
& 64.47 & {\scriptsize$\pm$0.26}
& 65.27 & {\scriptsize$\pm$0.67}
& \green{$\uparrow$~\small{14.01}} \\
& \(\mathtt{mhca}\)
& 68.90 & {\scriptsize$\pm$1.31}
& 56.41 & {\scriptsize$\pm$0.87}
& 63.77 & {\scriptsize$\pm$0.65}
& 63.03 & {\scriptsize$\pm$0.94}
& \green{$\uparrow$~\small{13.95}} \\
& \(\mathtt{protobin}\)
& 69.75 & {\scriptsize$\pm$0.45}
& {64.82} & {\scriptsize$\pm$0.36}
& {66.90} & {\scriptsize$\pm$0.26}
& {67.16} & {\scriptsize$\pm$0.36}
& \green{$\uparrow$~\small{13.82}} \\

\cmidrule(lr){1-11}

\multicolumn{2}{l}{\gray{$\mathtt{FT\!:\!scratch}$}}
& \gray{69.77} & {}
& \gray{69.19} & {}
& \gray{63.36} & {}
& \gray{67.44} & {}
& \\

\multicolumn{2}{l}{\gray{$\mathtt{FT\!:\!pretrained}$}}
& \gray{72.61} & {}
& \gray{83.43} & {}
& \gray{67.10} & {}
& \gray{74.38} & {}
& \\

\bottomrule
\end{tabular}
\end{table}
\endgroup

\begingroup
\setlength{\tabcolsep}{4pt}
\begin{table}[htb]
\caption{Downstream performance of CAP across pre-trained microscopy imaging encoders. FT:scratch and FT:pretrained have been obtained from DiChaViT \cite{pham2024enhancing} and IC-ViT \cite{lian2025isolated}, respectively. We use \(\mathtt{protobin}\) as the pooling architecture.  \(\Delta_\mathtt{CAP}\) indicates improvement versus $\mathtt{JFE+JAP}$ over the same encoder.}
\label{tab:lp_results_encoders}
\centering
\small
\begin{tabular}{
ll
c @{ }l
c @{ }l
c @{ }l
c
}
\toprule
Meth. & Encoder
& \multicolumn{2}{c}{CHAMMI}
& \multicolumn{2}{c}{JUMP}
& \multicolumn{2}{c}{Avg.}
& \(\Delta_{\mathtt{CAP}}\) \\
\midrule

\multirow{3}{*}{\(\mathtt{JFE+JAP}\)}
& IC-ViT 
& {56.41} & {\scriptsize$\pm$0.92}
& {39.34} & {\scriptsize$\pm$0.24} &  {47.88} & {\scriptsize$\pm$0.58} & \\
& OpenPhenom
& 58.88 & {\scriptsize$\pm$0.86}
& 33.83 & {\scriptsize$\pm$0.12}
& 46.36 & {\scriptsize$\pm$0.49}
& \\
& HPA-SC 
& 62.49 &  {\scriptsize$\pm$0.67} 
& 36.80 &  {\scriptsize$\pm$0.18} 
& 49.65 &  {\scriptsize$\pm$0.43} 
& \\

\cmidrule(lr){1-9}

\multirow{3}{*}{\(\mathtt{CAP}\)}
& IC-ViT
& 69.75 & {\scriptsize$\pm$0.45}
& {64.82} & {\scriptsize$\pm$0.36}
& 67.28 & {\scriptsize$\pm$0.41}
& \green{$\uparrow$~\small{19.41}} \\
& OpenPhenom
& 61.77 & {\scriptsize$\pm$0.87} 
& 71.44 & {\scriptsize$\pm$0.18}
& 66.61 & {\scriptsize$\pm$0.53}
& \green{$\uparrow$~\small{20.25}} \\
& HPA-SC 
& 75.10 &  {\scriptsize$\pm$0.68} 
& 52.69 &  {\scriptsize$\pm$0.16} 
& 63.90 &  {\scriptsize$\pm$0.42} 
& \green{$\uparrow$~\small{14.25}} \\

\cmidrule(lr){1-9}

\multirow{2}{*}{\gray{$\mathtt{FT}$}}  & \gray{scratch}
& \gray{69.77} & {}
& \gray{69.19} & {}
& \gray{69.48} & {}
& \\

 & \gray{pretrained} 
& \gray{72.61} & {}
& \gray{83.43} & {}
& \gray{78.02} & {}
& \\

\bottomrule
\end{tabular}
\end{table}
\endgroup

\section{Results}

\subsection{Downstream performance on MCI benchmarks}

\subsubsection{Performance of CAP across pooling architectures.} 
\tref{tab:lp_results} shows the downstream performance of CAP versus the standard probing baseline (JFE+JAP) on MCI datasets, for different pooling architectures. Here, we observe that CAP consistently outperforms the standard under all pooling architectures. More notably, we observe that the best performing architecture under CAP yields results much closer to full fine-tuning (67.16\% with 7\% gap to FT) versus the best performing architecture under the probing baseline (53.33\% with 21\% gap to FT). 

\subsubsection{Performance of CAP across pre-trained encoders.} 
\tref{tab:lp_results_encoders} shows the downstream performance of CAP versus the standard probing baseline (JFE+JAP) on MCI datasets, across fixed representations obtained from different microscopy imaging pre-trained encoders. Similar to \tref{tab:lp_results}, we observe that CAP consistently outperforms the standard under all pre-trained encoder settings.

\begin{figure*}[t]
    \centering
    \includegraphics[width=0.55\linewidth]{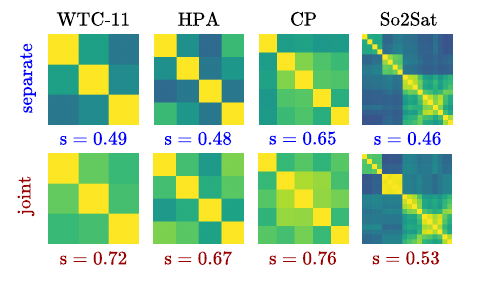}
\caption{Comparison of inter-channel feature diversity between separate and joint encoding. `\(\mathbf{s=}\)' denotes the average inter-channel cosine similarity for a given instance, averaged over 1000 random instances. For microscopy datasets (WTC, HPA, CP), we omit 75\% most similar patch tokens to remove effects of background similarity.}
\label{fig:interchan_b}
\end{figure*}

\begin{figure}[h]
\centering
\includegraphics[width=0.9\linewidth]{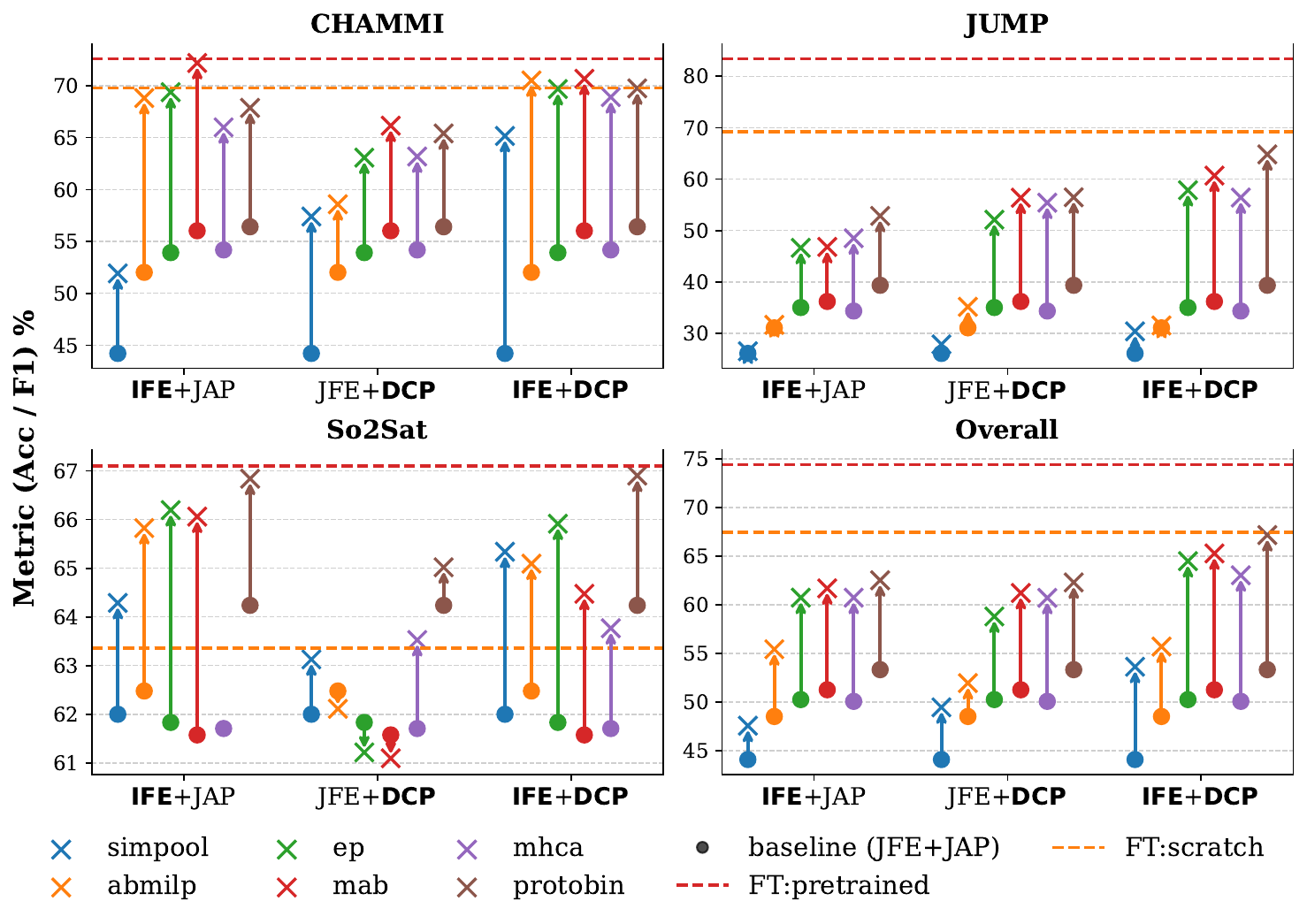}
\caption{Performance delta obtained by Independent Feature Encoding (IFE), Decoupled Pooling (DCP), and Channel-Aware Probing (IFE+DCP) over the probing baseline (JFE+JAP), for each dataset and aggregator. Results for FT:pretrained and FT:scratch are from IC-ViT \cite{lian2025isolated} and DiChaViT \cite{pham2024enhancing}, respectively. }
\label{fig:ablation}
\end{figure}

\subsection{Effects of IFE and DCP over the baseline (JFE+JAP)}
\label{sec:analysis}

\subsubsection{Independent Feature Encoding.} 

We claim in \sref{cap}, that when Joint Feature Encoding (JFE) is performed as in \eref{eq:jfe}, channel diversity is not preserved, and this degrades downstream performance. To first verify that channel-diversity is less preserved using JFE, we compare the fixed output representations of JFE and IFE. Specifically, for each instance in an MCI dataset, we perform forward passes with JFE and IFE, and obtain their respective feature maps. For each setting, we then compute the cosine similarity between tokens at a given spatial position, and obtain the average cosine similarity over all spatial positions, thus measuring the overall diversity between channels. \fref{fig:interchan_b} shows that IFE yields higher inter-channel feature diversity than JFE. We observe that this translates to increased downstream performance on all datasets over all pooling architectures, as shown in \fref{fig:ablation}. We attribute this to the rich channel feature diversity by IFE, resulting in better input features to the pooling architecture.



\subsubsection{Decoupled Pooling.}
\fref{fig:ablation} shows that, overall, DCP consistently improves over the standard JAP across pooling architectures. While Independent Feature Encoding (IFE) improves performance across all datasets, we observe a dataset-dependent trade-off between JAP, which treats features as a unified set, and DCP, which preserves channel-specific structure. We suspect that this trade-off may depend on how semantically distinct the channels are: microscopy datasets such as JUMP-CP and CHAMMI exhibit strongly semantically differentiated channels, whereas So2Sat channels correspond to wavelength bands with weaker semantic separation. We aim to investigate how channel semantics influence pooling strategies in future work.








\subsection{Computational complexity of CAP}
\label{sec:compute}

As the compute cost of ViTs scales quadratically with sequence length, the FLOPs for obtaining pre-computed features using IFE, which increases linearly with channels and quadratic with sequence length (\(O(CN^2)\)) remains theoretically much lesser than JFE which increases quadratically for both channel and sequence length (\(O(C^2N^2)\)). Therefore, we do not perform further analysis into the cost associated with feature extraction. 

\begin{figure}[h]
\centering
\begin{minipage}[c]{0.4\linewidth}
    \centering
        \begin{subfigure}[c]{\linewidth}
        \centering
        \includegraphics[width=\linewidth, page=1]{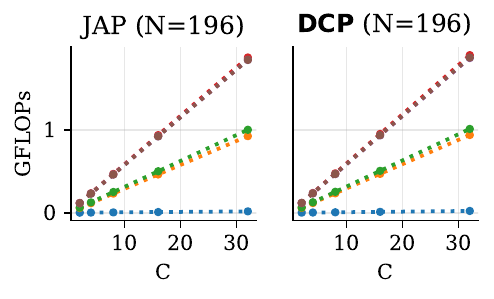}
    \subcaption{}
    \label{fig:cost_a}
    \end{subfigure}
\end{minipage}%
\begin{minipage}[c]{0.44\linewidth}
    \centering
    \begin{subfigure}[c]{\linewidth}
        \centering
        \includegraphics[width=\linewidth, page=2]{imgs/compute.pdf}
        \subcaption{}
        \label{fig:cost_b}
    \end{subfigure}
\end{minipage}
\begin{minipage}[c]{0.14\linewidth}
    \centering
    \begin{subfigure}[c]{\linewidth}
        \centering
        \includegraphics[width=\linewidth, page=3]{imgs/compute.pdf}
    
    \end{subfigure}
\end{minipage}
\caption{JAP vs. DCP FLOP count with (a) fixed \(N\) and varying \(C\), and (b) varying  \(N\) and fixed \(C\),  for each aggregator.}
\end{figure}

On the compute complexity of DCP, we mention in \sref{cap} that the overall increase in FLOPs for DCP over JAP is negligible due to the second forward-pass in \eref{dcp_fwd} being an aggregation in the channel (\(C\)) dimension, where \(C \ll N\). 
As shown in  \fref{fig:cost_a} and \fref{fig:cost_b}, a near-identical computational cost is observed between DCP and JAP, regardless of whether the number of channels (\(C\)) or the number of tokens (\(N\)) is varied. Although DCP contains two forward passes, it is computationally equivalent to the probing baseline, and the performance increase obtained by DCP largely outweighs the minor (relatively unobservable) increase in FLOPs.

\section{Conclusion}

We propose Channel-Aware Probing, an efficient approach for improving downstream probing performance on multi-channel imaging (MCI) datasets. Our method explicitly leverages the inherent inter-channel diversity present in MCI data through two complementary components: Independent Feature Encoding and Decoupled Pooling. Across the CHAMMI, JUMP-CP, and So2Sat benchmark datasets, each component independently improves performance over the standard probing baseline by approximately 9\%. When combined into Channel-Aware Probing, these gains increase to 14\%. As a result, our approach substantially narrows the gap between probing-based evaluations and full fine-tuning in MCI settings, establishing a stronger baseline for probing on multi-channel imaging datasets.


\newpage


\begin{thebibliography}{10}
\providecommand{\url}[1]{\texttt{#1}}
\providecommand{\urlprefix}{URL }
\providecommand{\doi}[1]{https://doi.org/#1}

\bibitem{assran2025v}
Assran, M., Bardes, A., Fan, D., et~al.: V-jepa 2: Self-supervised video models enable understanding, prediction and planning. arXiv preprint arXiv:2506.09985  (2025)

\bibitem{bao2023channel}
Bao, Y., Sivanandan, S., Karaletsos, T.: Channel vision transformers: An image is worth 1 x 16 x 16 words. In: The Twelfth International Conference on Learning Representations (2024)

\bibitem{bourriez2024chada}
Bourriez, N., Bendidi, I., Cohen, E., Watkinson, G., Sanchez, M., Bollot, G., Genovesio, A.: Chada-vit: Channel adaptive attention for joint representation learning of heterogeneous microscopy images. In: Proceedings of the IEEE/CVF Conference on Computer Vision and Pattern Recognition. pp. 11556--11565 (2024)

\bibitem{bray2016cell}
Bray, M.A., Singh, S., Han, H., Davis, C.T., Borgeson, B., Hartland, C., Kost-Alimova, M., Gustafsdottir, S.M., Gibson, C.C., Carpenter, A.E.: Cell painting, a high-content image-based assay for morphological profiling using multiplexed fluorescent dyes. Nature protocols  \textbf{11}(9),  1757--1774 (2016)

\bibitem{chandrasekaran2023jump}
Chandrasekaran, S.N., Ackerman, J., Alix, E., Ando, D.M., Arevalo, J., Bennion, M., Boisseau, N., Borowa, A., Boyd, J.D., Brino, L., et~al.: Jump cell painting dataset: morphological impact of 136,000 chemical and genetic perturbations. BioRxiv pp. 2023--03 (2023)

\bibitem{chandrasekaran2024three}
Chandrasekaran, S.N., Cimini, B.A., Goodale, A., Miller, L., Kost-Alimova, M., Jamali, N., Doench, J.G., Fritchman, B., Skepner, A., Melanson, M., et~al.: Three million images and morphological profiles of cells treated with matched chemical and genetic perturbations. Nature Methods  \textbf{21}(6),  1114--1121 (2024)

\bibitem{chen2023chammi}
Chen, Z.S., Pham, C., Wang, S., Doron, M., Moshkov, N., Plummer, B., Caicedo, J.C.: Chammi: A benchmark for channel-adaptive models in microscopy imaging. Advances in Neural Information Processing Systems  \textbf{36},  19700--19713 (2023)

\bibitem{doron2023unbiased}
Doron, M., Moutakanni, T., Chen, Z.S., Moshkov, N., Caron, M., Touvron, H., Bojanowski, P., Pernice, W.M., Caicedo, J.C.: Unbiased single-cell morphology with self-supervised vision transformers. bioRxiv  (2023)

\bibitem{dosovitskiy2020image}
Dosovitskiy, A., Beyer, L., Kolesnikov, A., Weissenborn, D., Zhai, X., Unterthiner, T., Dehghani, M., Minderer, M., Heigold, G., Gelly, S., Uszkoreit, J., Houlsby, N.: An image is worth 16x16 words: Transformers for image recognition at scale. In: International Conference on Learning Representations (2021)

\bibitem{everingham2010pascal}
Everingham, M., Van~Gool, L., Williams, C.K., Winn, J., Zisserman, A.: The pascal visual object classes (voc) challenge. International journal of computer vision  \textbf{88}(2),  303--338 (2010)

\bibitem{gupta2024subcell}
Gupta, A., Wefers, Z., Kahnert, K., Hansen, J.N., Leineweber, W., Cesnik, A., Lu, D., Axelsson, U., Ballllosera~Navarro, F., Karaletsos, T., et~al.: Subcell: Vision foundation models for microscopy capture single-cell biology. bioRxiv pp. 2024--12 (2024)

\bibitem{hense2024xmil}
Hense, J., Jamshidi~Idaji, M., Eberle, O., Schnake, T., et~al.: Xmil: Insightful explanations for multiple instance learning in histopathology. Advances in Neural Information Processing Systems  \textbf{37},  8300--8328 (2024)

\bibitem{9672063}
Herruzo, P., Gruca, A., Lliso, L., Calbet, X., Rípodas, P., Hochreiter, S., Kopp, M., Kreil, D.P.: High-resolution multi-channel weather forecasting – first insights on transfer learning from the weather4cast competitions 2021. In: 2021 IEEE International Conference on Big Data (Big Data). pp. 5750--5757 (2021)

\bibitem{lee2019set}
Lee, J., Lee, Y., Kim, J., Kosiorek, A., Choi, S., Teh, Y.W.: Set transformer: A framework for attention-based permutation-invariant neural networks. In: International conference on machine learning. pp. 3744--3753. PMLR (2019)

\bibitem{lian2025isolated}
Lian, W., Lindblad, J., Micke, P., Sladoje, N.: Isolated channel vision transformers: From single-channel pretraining to multi-channel finetuning. arXiv preprint arXiv:2503.09826  (2025)

\bibitem{nguyen2023climax}
Nguyen, T., Brandstetter, J., Kapoor, A., Gupta, J.K., Grover, A.: Climax: A foundation model for weather and climate. arXiv preprint arXiv:2301.10343  (2023)

\bibitem{oquab2023dinov2}
Oquab, M., Darcet, T., Moutakanni, T., Vo, H., Szafraniec, M., Khalidov, V., Fernandez, P., Haziza, D., Massa, F., El-Nouby, A., et~al.: Dinov2: Learning robust visual features without supervision. arXiv preprint arXiv:2304.07193  (2023)

\bibitem{pham2025cha}
Pham, C., Caicedo, J.C., Plummer, B.A.: Cha-maevit: unifying channel-aware masked autoencoders and multi-channel vision transformers for improved cross-channel learning. arXiv preprint arXiv:2503.19331  (2025)

\bibitem{pham2024enhancing}
Pham, C., Plummer, B.: Enhancing feature diversity boosts channel-adaptive vision transformers. Advances in Neural Information Processing Systems  \textbf{37},  89782--89805 (2024)

\bibitem{przewikezlikowski2025beyond}
Przewi{\k{e}}{\'z}likowski, M., Balestriero, R., Jasi{\'n}ski, W., {\'S}mieja, M., Zieli{\'n}ski, B.: Beyond [cls]: Exploring the true potential of masked image modeling representations. In: Proceedings of the IEEE/CVF International Conference on Computer Vision. pp. 23442--23452 (2025)

\bibitem{psomas2025attention}
Psomas, B., Christopoulos, D., Baltzi, E., et~al.: Attention, please! revisiting attentive probing for masked image modeling. arXiv preprint arXiv:2506.10178  (2025)

\bibitem{psomas2023keep}
Psomas, B., Kakogeorgiou, I., Karantzalos, K., Avrithis, Y.: Keep it simpool: Who said supervised transformers suffer from attention deficit? In: Proceedings of the IEEE/CVF International Conference on Computer Vision. pp. 5350--5360 (2023)

\bibitem{rauch2025unmute}
Rauch, L., Heinrich, R., Ghaffari, H., Miklautz, L., Moummad, I., Sick, B., Scholz, C.: Unmute the patch tokens: Rethinking probing in multi-label audio classification. arXiv preprint arXiv:2509.24901  (2025)

\bibitem{rauch2025can}
Rauch, L., Heinrich, R., Moummad, I., Joly, A., Sick, B., Scholz, C.: Can masked autoencoders also listen to birds? arXiv preprint arXiv:2504.12880  (2025)

\bibitem{shao2025mil}
Shao, D., Chen, R.J., Song, A.H., et~al.: Do mil models transfer? arXiv preprint arXiv:2506.09022  (2025)

\bibitem{thul2017subcellular}
Thul, P.J., {\AA}kesson, L., Wiking, M., Mahdessian, D., Geladaki, A., Ait~Blal, H., Alm, T., Asplund, A., Bj{\"o}rk, L., Breckels, L.M., et~al.: A subcellular map of the human proteome. Science  \textbf{356}(6340),  eaal3321 (2017)

\bibitem{viana2023integrated}
Viana, M.P., Chen, J., Knijnenburg, T.A., Vasan, R., Yan, C., Arakaki, J.E., Bailey, M., Berry, B., Borensztejn, A., Brown, E.M., et~al.: Integrated intracellular organization and its variations in human ips cells. Nature  \textbf{613}(7943),  345--354 (2023)

\bibitem{hpa-single-cell-image-classification}
Winsnes, C., Lundberg, E., Maggie, Culliton, P., Le, T., UAxelsson, Ouyang, W.: Human protein atlas - single cell classification. \url{https://kaggle.com/competitions/hpa-single-cell-image-classification} (2021), kaggle

\bibitem{xun2023microsnoop}
Xun, D., Wang, R., Zhang, X., Wang, Y.: Microsnoop: A generalist tool for microscopy image representation. The Innovation  \textbf{5}(1),  100541 (2024). \doi{https://doi.org/10.1016/j.xinn.2023.100541}

\bibitem{zhou2021ibot}
Zhou, J., Wei, C., Wang, H., Shen, W., Xie, C., Yuille, A., Kong, T.: ibot: Image bert pre-training with online tokenizer. arXiv preprint arXiv:2111.07832  (2021)

\bibitem{so2sat}
Zhu, X.X., Hu, J., Qiu, C., Shi, Y., Kang, J., Mou, L., Bagheri, H., Häberle, M., Hua, Y., Huang, R., Hughes, L., Li, H., Sun, Y., Zhang, G., Han, S., Schmitt, M., Wang, Y.: So2sat lcz42: A benchmark dataset for global local climate zones classification (2019), \url{https://arxiv.org/abs/1912.12171}

\end{thebibliography}

\end{document}